\documentclass[conference]{IEEEtran}
\IEEEoverridecommandlockouts
\usepackage{cite}
\usepackage{amsmath,amssymb,amsfonts}
\usepackage{multirow}
\usepackage{algorithm}
\usepackage{algpseudocode}
\usepackage{balance}
\usepackage{todonotes}
\usepackage{graphicx}
\usepackage{textcomp}
\usepackage{xcolor}
\def\BibTeX{{\rm B\kern-.05em{\sc i\kern-.025em b}\kern-.08em
    T\kern-.1667em\lower.7ex\hbox{E}\kern-.125emX}}
\begin{document}

\title{Zero-Shot Meta-Learning for Small-Scale Data From Human Subjects}

\author{\IEEEauthorblockN{
    Julie Jiang\IEEEauthorrefmark{1}\IEEEauthorrefmark{2}, 
    Kristina Lerman\IEEEauthorrefmark{1}\IEEEauthorrefmark{2}, 
    Emilio Ferrara\IEEEauthorrefmark{1}\IEEEauthorrefmark{2}\IEEEauthorrefmark{3}
    \IEEEauthorblockA{juliej@isi.edu, lerman@isi.edu, emiliofe@usc.edu}}
    \IEEEauthorblockA{
        \IEEEauthorrefmark{1}
        USC Information Sciences Institute, Marina del Rey, CA}
    \IEEEauthorblockA{
        \IEEEauthorrefmark{2}
        USC Viterbi School of Engineering, Los Angeles, CA}
    \IEEEauthorblockA{
        \IEEEauthorrefmark{3}
        USC Annenberg School of Communication, Los Angeles, CA}
    }

\maketitle



\begin{abstract}
While developments in machine learning led to impressive performance gains on big data, many human subjects data are, in actuality, small and sparsely labeled. Existing methods applied to such data often do not easily generalize to out-of-sample subjects. Instead, models must make predictions on test data that may be drawn from a different distribution, a problem known as \textit{zero-shot learning}. To address this challenge, we develop an end-to-end framework using a meta-learning approach, which enables the model to rapidly adapt to a new prediction task with limited training data for out-of-sample test data. We use three real-world small-scale human subjects datasets (two randomized control studies and one observational study), for which we predict treatment outcomes for held-out treatment groups. Our model learns the latent treatment effects of each intervention and, by design, can naturally handle multi-task predictions. We show that our model performs the best holistically for each held-out group and especially when the test group is distinctly different from the training group. Our model has implications for improved generalization of small-size human studies to the wider population.

\end{abstract}
\maketitle

\section{Introduction}

Human subjects studies are critical to our understanding of human health, psychology, and decision-making as well as the effectiveness of interventions to improve individual well-being~\cite{ildstad2001small,hamburg2010path}. Though such studies remain the gold standard of scientific discovery \cite{ildstad2001small,day2018recommendations}, many are small and sparsely labeled due to regulatory challenges, ethical considerations \cite{hackshaw2008small}, data availability (e.g., investigating rare diseases \cite{day2018recommendations}),  recruitment costs, and data collection \cite{martin2017much}. Another challenge is generalizing the results of human subjects experiments to real-world settings since the inclusion-exclusion criteria (criteria that specify whether a participant is eligible to participate in a study) of these studies skew the study population compared to the general population~\cite{davis1994generalizing,flather2006generalizing,hong2017generalizing}.
As such, there is a need for developing accurate predictive models from data collected from small-scale human subjects studies. For example, these models can help make optimal treatment decisions based on an individual's predicted treatment outcomes \cite{hamburg2010path}. However, most efforts in this area have been constrained by the limited availability of labeled data and the lack of generalizability to out-of-sample populations. 

In recent years, deep learning methods have largely benefited from the availability of large-scale labeled data in areas such as computer vision \cite{voulodimos2018deep} and natural language processing \cite{devlin2018bert}. However, these methods have had limited success in predicting outcomes in human subject studies \cite{rajkomar2018scalable}, where many researchers continue to favor traditional methods such as Elastic Nets or Random Forests \cite{chekroud2016cross, walsh2017predicting} instead of deep learning architectures, which generally necessitate a large amount of labeled data yet have limited capacity for transferring knowledge \cite{sun2017revisiting,marcus2018deep}, hindering their ability to generalize to complex yet small human subjects datasets and tasks \cite{shah2019artificial}. Transfer learning is a new approach that could address this challenge. For example, transfer learning of deep architectures was applied to the survival analysis of cancer patients by leveraging other cancer datasets \cite{li2016transfer,qiu2020meta}.

\begin{figure}
    \centering
    
    \label{fig:motivation}
    \includegraphics[width=\linewidth]{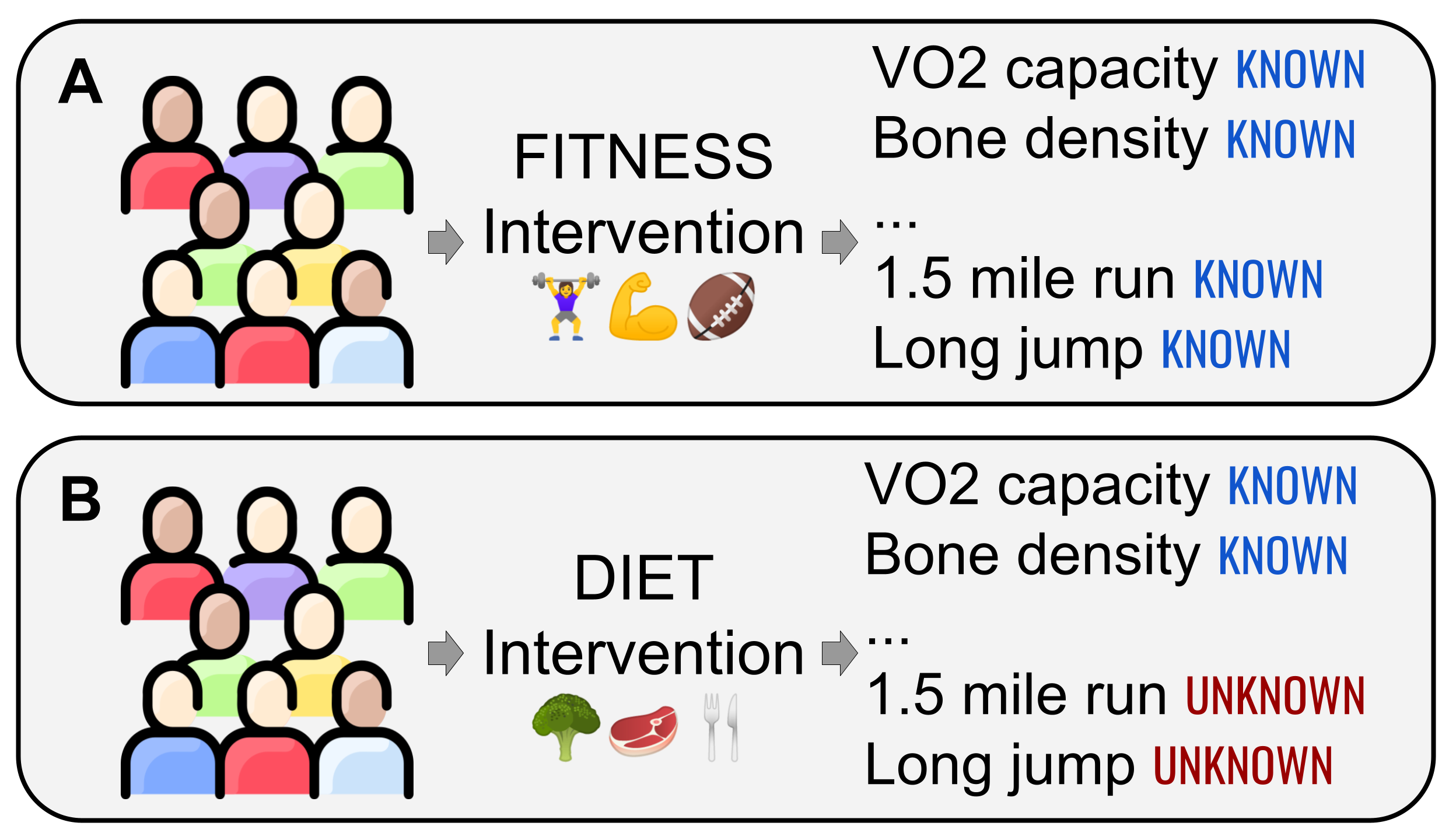}
    \caption{A motivating example: given two treatment groups of participants and a set of known assessment outcomes, we want to predict unknown assessment outcomes (e.g., \textit{1.5 mile run}, \textit{long jump}) for one treatment group by transferring from other assessment outcomes.}
    
\end{figure}

Inspired by transfer learning approaches, we introduce a more general paradigm for such human subject prediction problems based on meta-learning. We frame these tasks as zero-shot learning (ZSL) problems, where we make predictions for the outcomes of treatment for subjects (or \textit{shots}) in a test treatment group without being trained on any (\textit{zero}) subject in that group for the target tasks. Guided by the paradigm of meta-learning, or \textit{learning to learn}, our method can rapidly generalize to unseen groups \cite{finn2017model}.   Unlike other ZSL approaches, we do not require human-annotated attributes to characterize the treatment groups \cite{romera2015embarrassingly}.  Moreover, our method is inherently capable of making multi-task predictions, since meta-learning transfers knowledge across tasks. We emphasize that our problem is distinct from most literature on ZSL, which specifically addresses zero-shot classification in which the test set contains unseen labels from the training set \cite{romera2015embarrassingly,verma2020meta,xian2016latent,felix2018multi}, as well as from most literature on meta-learning \cite{finn2017model, nichol2018reptile,li2016transfer}, which enables quick adaptation in a few-shot (rather than zero-shot) scenario. 


As a motivation, consider a study where participants are randomly assigned to an intervention: either a \textit{diet} regimen or a \textit{fitness} training treatment (illustrated in Fig. \ref{fig:motivation}). After the intervention, the \textit{fitness} group is given a fitness assessment but not the \textit{diet} group. Suppose both treatments have a substantial, but unknown, effect on the fitness capabilities of both groups of people, altering the underlying distribution of potential fitness assessment results. If we were to predict the fitness test results of the \textit{diet} group, we might reasonably expect that, all else being equal, the \textit{diet} group will underperform in certain training assessments compared to the \textit{fitness} group (e.g., endurance tests). However, traditional machine learning models, having never seen how the \textit{diet} group participants perform on the fitness tests, can only make predictions based on the assumption that they too underwent a \textit{fitness} treatment plan. We address this problem by meta-learning how subjects from both treatment groups respond to tasks similar to fitness assessment. Our method learns treatment representations that capture the latent effects of treatments to overcome the challenges of ZSL. Our method is flexible and can be generalized to predict several target values (i.e., multitask prediction).

In sum, the contributions of this paper are as follows:
\begin{itemize}
    \item We propose a meta-learning-based treatment outcome prediction for \textit{out-of-sample} users (users who underwent an unseen treatment plan) without having domain knowledge or characterization of the treatment plan.
    \item We demonstrate our model achieves superior performance on three real-world small-scale human subjects datasets (two controlled studies and one observational study). 
    \item Our model is resistant to overfitting, making it especially advantageous when the target variable distribution of the test data is most distinct from that of the training data.
\end{itemize}


\section{Related Work}
\subsection{Transfer Learning and Domain Adaptation}  In transfer learning, knowledge is transferred from the source data to the target data. Usually, the source data is larger and easier to acquire, whereas the target data is small and scarce. A transfer learning model can be trained on the source data and then fine-tuned on the target data. There are two main types of transfer learning: \textit{domain} adaptation and \textit{task} adaptation. In domain adaptation, the source and target domains are different but somewhat similar \cite{pan2010survey}. Transfer learning is widely studied in computer vision and language modeling \cite{csurka2017domain,ruder2019transfer}. For example, a language model may be trained on an English corpus, for which data is abundant, but tested on an Arabic corpus, for which data is scarce.

Compared to conventional methods, models designed for domain adaptation must learn from domain-invariant features in the training data. Importantly, it should not overfit the training data such that it compromises its generalizability to a different domain. Discrepancy-based methods aim to minimize the discrepancy between the training and the target domains \cite{long2017deep}. Adversarial approaches differentiate target domain data from the source domain data \cite{tzeng2017adversarial}. In an application to predict anti-cancer drug response, \cite{zhu2020ensemble} used an ensemble model to transfer knowledge across various drug response datasets. Transfer learning has also been shown to improve survival analysis in real-world high-dimensional cancer datasets\cite{li2016transfer}.

More challenging than domain adaptation is the problem of task adaptation. There is substantive interest in developing general-purpose, ubiquitous models that are not task-specific \cite{caruana1997multitask,sousa2006task,zhang2017survey,liu2019multitask}. A transfer learning approach to these problems transfers knowledge learned from a set of tasks to a different set of tasks that are similar in some ways \cite{torrey2010transfer}. While, such methods have been widely explored in domains such as reinforcement learning \cite{taylor2009transfer}, natural language processing \cite{yogatama2019learning,vu2020exploring}, and computer vision \cite{li2020transfer}, their applications to human behavioral datasets remain sparse.

\subsection{Zero-/Few-shot Learning} Few-shot learning (FSL) refers to a type of transfer learning problem in which the training data is very limited \cite{wang2020generalizing}. Compared to traditional machine learning models, FSL models must generalize rapidly from a very small amount of data without overfitting. FSL has garnered lots of attention in many real-life applications in which data is limited. A common application of FSL is few-shot classification: the classifier is trained with only $k$ (a small number) examples each from $n$ target classes before being deployed on the test set. Similarly, a $k$-shot regressor is only trained on  $k$ training examples. When $k=0$, FSL becomes zero-shot learning (ZSL). ZSL/FSL models typically require pre-training on a larger source data, followed by transferring the knowledge to the target data. To enable ZSL, the model must acquire knowledge from other modalities, such as through auxiliary attributes available for both source and target data \cite{romera2015embarrassingly}. ZSL/FSL are naturally well-positioned to tackle multi-task learning \cite{caruana1997multitask,zhang2017survey} or task adaptation \cite{ravi2016optimization,finn2017model} by transferring knowledge from source tasks that have large training examples to target tasks that have limited or no training examples. Many ZSL/FSL methods, including ours, are solved via meta-learning, which we introduce in detail next.

\subsection{Meta-learning} Meta-learning is a new type of machine learning paradigm that \textit{learns to learn}. It is an end-to-end framework that can learn new knowledge with very few training examples, thereby enabling ZSL/FSL. Matching Networks \cite{vinyals2016matching} or Prototypical Networks \cite{snell2017prototypical} learn representations of data points and leverage the similarity of representations between any two data points to enable zero- or few-shot classification. \cite{santoro2016meta} proposed memory-augmented neural networks to store previous training examples for future use. 

More recently, optimization-based methods were developed to separate the meta-learning framework into the base-learner and the meta-learner. The base-learner is trained normally, while the meta-learner improves the base-learner. \cite{ravi2016optimization} introduced the LSTM Meta-learner, which meta-learns an exact optimization rule to update the base-learner's parameters for multi-task few-shot classification. On the other hand, Model-Agnostic Meta-Learning (MAML)  is a more generalized approach to multi-task learning for supervised and reinforcement learning \cite{finn2017model}.  MAML meta-learns a set of optimal initialization weights for the base-learner to allow it to generalize quickly to any task \cite{finn2017model}. Reptile \cite{nichol2018reptile}, a first-order MAML, was developed as an approximation of MAML, vastly reducing the computational overhead while maintaining comparable performance. Another model similar to MAML is Meta-SGD, which learns not only good initialization weights but also optimal update directions and learning rates \cite{li2017meta}. However, all of these flavors of meta-learning models are designed for FSL for a new task rather than ZSL for a new domain.

While meta-learning variants have been developed for ZSL, they generally require human-annotated class attributes that describe seen and unseen classes \cite{verma2020meta,xian2016latent,felix2018multi}. We propose a meta-learning framework that enables ZSL for both regression and classification problems, without the need for human-annotated class attributes.

\subsubsection{Distinction From Our Work}

While our method borrows intuitions from meta-learning literature, to the best of our knowledge, these existing models do not natively apply to our problem setting. First of all, most methods only concern zero-shot or few-shot classification (e.g. detecting a new category of images or samples that are unseen during training) \cite{romera2015embarrassingly,verma2020meta,xian2016latent,felix2018multi}. In contrast, we are trying to achieve regression outcome predictions for unseen users from a different treatment group.  Other meta-learning methods capable of prediction regression outcomes address fast adaptation to new tasks via few-shot learning \cite{finn2017model, nichol2018reptile,li2016transfer}. Our task is a zero-shot problem given we have no target data for the out-of-sample users.  These problems are characteristically different from ours, and thus cannot be directly compared.

Meta-learning as is used in this work is also not to be confused with the \textit{meta-learners} proposed by \cite{kunzel2019metalearners} for causal inference (e.g., X-learner, T-learner, and S-learner). The meta-learners by \cite{kunzel2019metalearners} estimates heterogeneous treatment effects based on data available from \textit{all} treatment groups. This problem setting is distinctly different from ours, as the predictions we are making are the treatment outcomes for a separate treatment group and this data is held-out during training.

\section{Methods}
\subsection{Preliminaries}
We first formally define our problem and introduce some notations. Given human subject data that is separated into disjoint treatment groups, our problem is (multi-task) prediction for data points for a set of held-out treatment groups. The dataset consists of  $i=1...N$ individuals, each described by a feature vector $\textbf x^i$, a treatment group indicator $g^i$, and a set of target variables $\{\textbf y_\tau^i|\tau\in T_{\text{target}}\}$, one for each task $T_{\text{target}}$. Suppose $g^*$ represents the held-out group, then for each task $\tau\in T_{\text{target}}$, the training $D_\tau^{\text{train}} = \{(\textbf x^i,g^i,\textbf y_\tau^i) | g^i\ne g^*\}$ and testing $D_\tau^{\text{test}} = \{(\textbf x^i,g^i,\textbf y^i_\tau)|g^i=g^*\}$ data are distinctly partitioned by $g^*$. The target variables $\{\textbf y_\tau^i|\tau\in T_{\text{target}},g^i=g*\}$ are unknown.

To address this zero-shot problem, we develop a meta-learner that can learn the latent effects of each treatment on each group of individuals and good initialization weights for $T_{\text{target}}$ through learning to predict training tasks $T_{\text{training}}$. For any task $\tau\in T_{\text{training}}$, there is a training dataset $D_\tau^{\text{train}} = \{(\textbf x^i,g^i,\textbf y_\tau^i) | g^i\ne g^*\}$ and a fine-tuning dataset $D_\tau^{\text{fine-tune}} = \{(\textbf x^i,g^i,\textbf y^i_\tau)|g^i=g^*\}$ that are also distinctly partitioned by $g^*$, except all target variables are known.

In the following sections, we first describe how we find training tasks $T_{\text{training}}$ for meta-training. Then, we detail the model architecture of the base-learner that learns the treatment representations. Finally, we discuss the meta-training and meta-testing procedures.

\subsection{Training Tasks} In controlled human subject trials, features of individual participants are collected either pre-treatment, during the treatment, or post-treatment. Assuming that the treatment has a consistent predictable effect on each individual, we can infer the treatment effects using any post-treatment features. Therefore, we can select any feature variables that are collected post-treatment as the target variables to predict in the training tasks, while using pre-treatment and during-the-treatment features as input features.

We outline several strategies to find such training tasks. The simplest strategy would be to use all available post-treatment training features as the training tasks, barring any post-treatment features not affected by the treatment (e.g., height taken both before and after treatment remains the same), so that the model does not simply memorize pre-treatment features. However, this depends on the quality of all post-treatment features and how much they are associated with the target tasks. 

As an improvement, we take inspiration from feature selection methods, which find features highly informative for the target tasks. To do so, we take all post-treatment training features and compare them with the targets $ Y^{\text{train}} =\{\textbf y_\tau^i|\tau\in T_{\text{target}},g^i\neq g^*\}$ to find the most informative features. We experiment with using Pearson correlation, where we take the features most correlated with the target variables (positive or negative). Alternatively, we can use Mutual Information \cite{kraskov2004estimating} to measure the pairwise dependency between any feature variable and target variable in the training set and take the features that have the highest mutual information score. Finally, we can use domain knowledge to manually craft features (post-treatment or not) that are highly relevant to the target tasks. We will not explore this last method as we aim to build a domain-agnostic model for individual treatment effect prediction.

\begin{figure}
    \centering
    \includegraphics[width=0.5\linewidth]{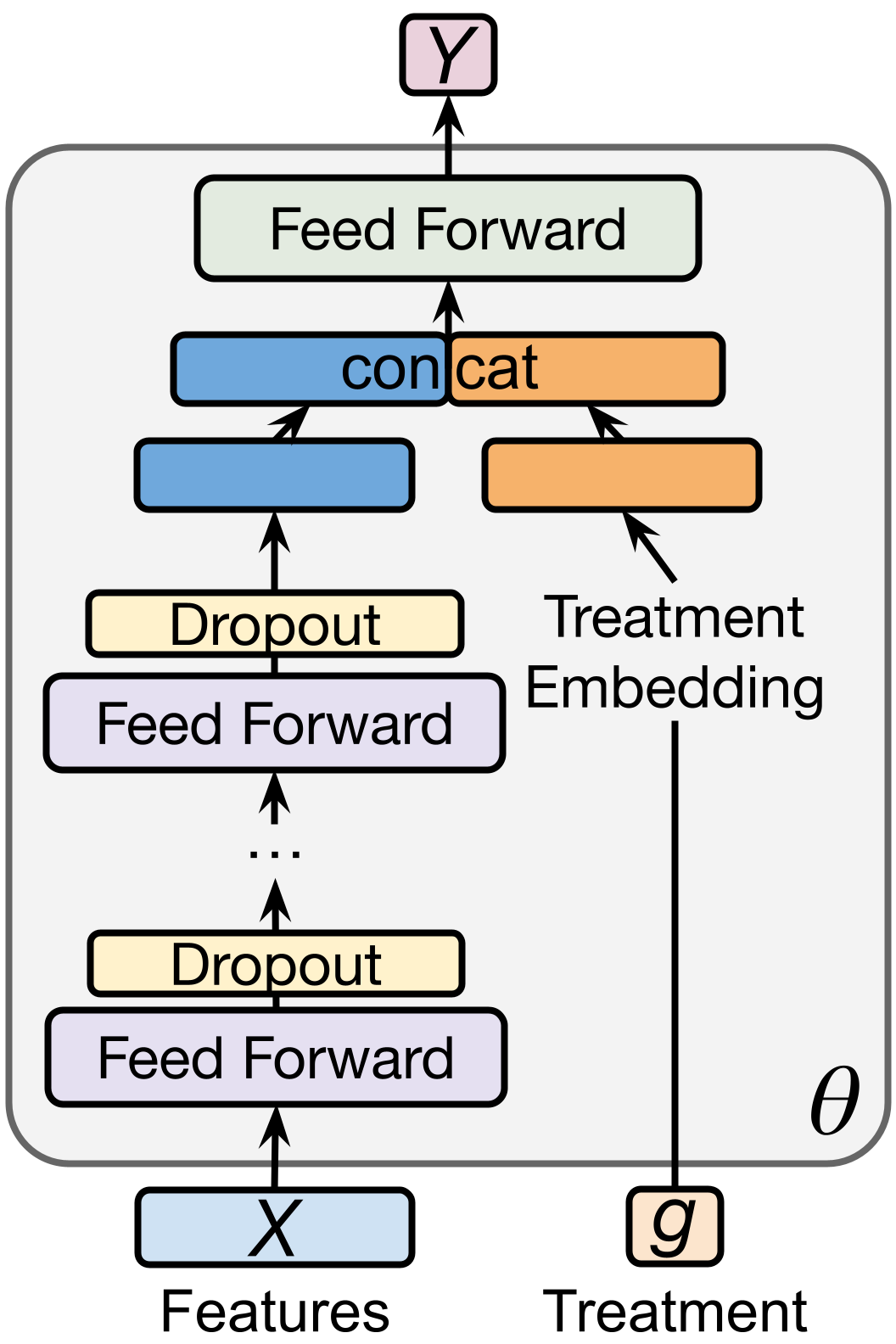}
    \caption{The model architecture of the base-learner, parameterized by $\theta$.}
    \label{fig:base_learner}
\end{figure}

\begin{algorithm}[tb]
\caption{Meta-training}
\label{alg:meta-training}
\begin{algorithmic}[1] 
\State Randomly initialize $\theta$
\While{not done}
    \State Sample task $\tau\in  T_{\text{train}}$ and data $\{ D_{\tau}^{\text{train}\prime},  D_{\tau}^{\text{test}\prime}\}$
    \State $\tilde \theta\gets U^{k}(\theta, { D_\tau^{\text{train}\prime}})$ \Comment{Train}
    \State $\tilde \theta\gets U^{k}(\tilde{\theta},  D_\tau^{\text{fine-tune}\prime})$ \Comment{Fine-tune}
    \State $\theta \gets \theta + \epsilon(\theta - \tilde{\theta}) $ \Comment{Update}
\EndWhile
\State \textbf{return} $\theta$
\end{algorithmic}
\end{algorithm}
\subsection{Base-learner} \label{sec:base-learner}
Fig.~\ref{fig:base_learner} illustrates the base-learner, which consists of (1) a feature extractor that is applied to features $\textbf x$ and (2) a treatment embedding layer that is applied to treatment group indicator $g$. All weights in the base-learner are parameterized by $\theta$. The feature extractor extracts a latent representation of the input features. It consists of a series of feed-forward layers, each optionally followed by a Dropout layer to avoid overfitting \cite{srivastava2014dropout}. The treatment embedding layer embeds the treatment groups by constructing a latent representation per group. The feature and treatment representations are concatenated and sent through a feed-forward layer, which outputs a predicted value. Following \cite{li2017meta}, we accelerate and stabilize training with weight normalization \cite{salimans2016weight}.

However, applying the base-learner directly to our problem is futile because the treatment embedding of the test treatment group will never be used during training and hence not learned. In the following section, we discuss how we leverage meta-learning to learn good treatment representations for all treatment groups, including the held-out test treatment group.

\subsection{Meta-learner} Our meta-learning method is based on Reptile \cite{nichol2018reptile}, a type of first-order approximation of MAML \cite{finn2017model}. Similar to MAML \cite{finn2017model}, Reptile aims to find a set of good initialization weights such that it can quickly adapt to any training task with minimal fine-tuning. A benefit of Reptile is that by computing only the first-order derivatives, it offers competitive performance at a lower computational cost \cite{nichol2018reptile}. As such, we derive our meta-learner from Reptile and extend it from few-shot to zero-shot learning to predict for unseen groups.

\begin{figure*}
    \centering
    \includegraphics[width=\linewidth]{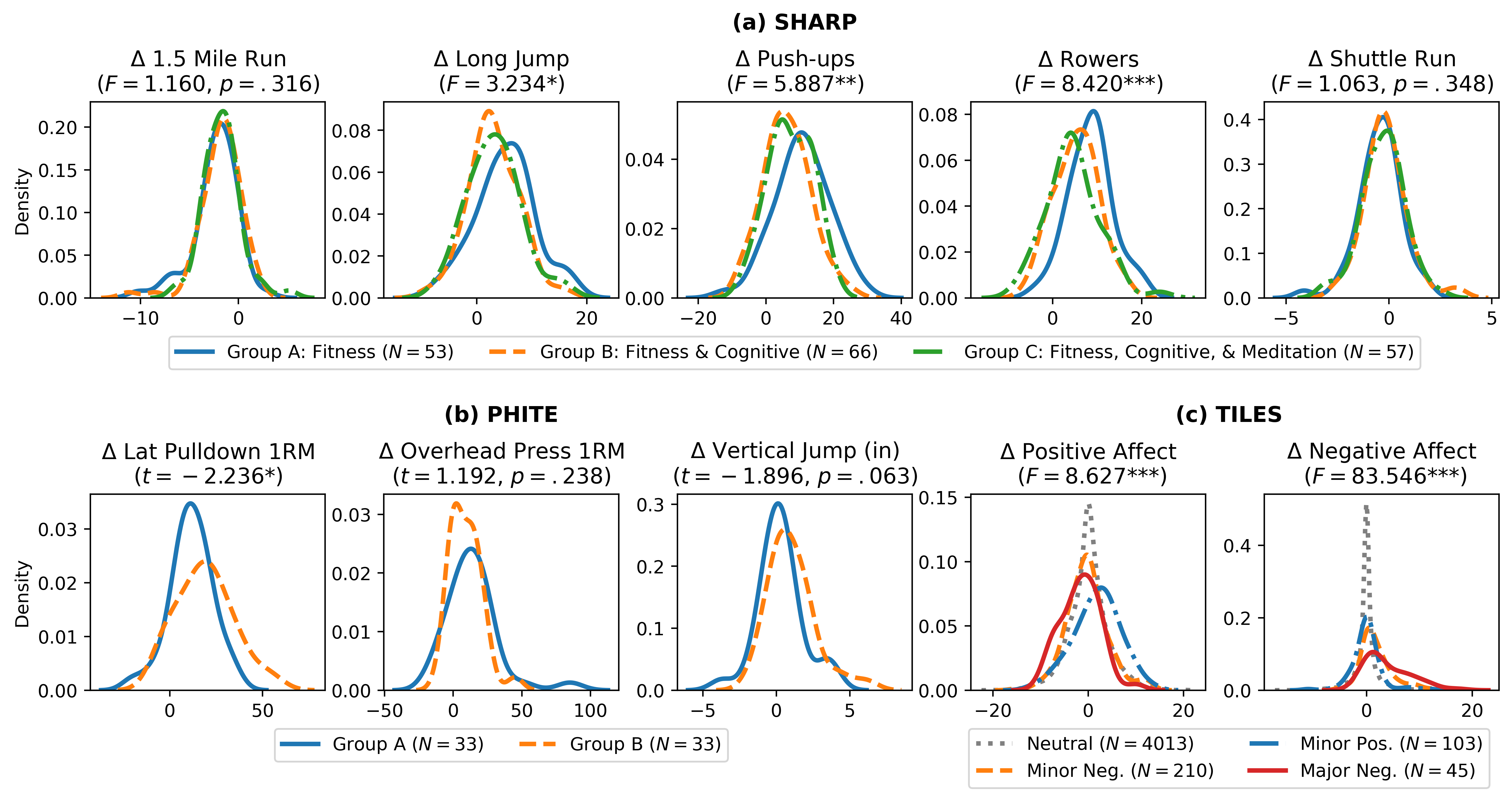}
    \caption{Distributions of the treatment outcomes of the change in (a) the fitness test scores of SHARP, (b) the fitness test scores of PHITE, and (c) the affect scores of TILES. We indicate whether the distributions are statistically significantly different between groups using ANOVA for SHARP and TILES (multigroup) and $t$-tests for PHITE (two groups), where *$ p<.05$, ** $ p<.01$, and *** $ p<.001$.}
    \label{fig:data_dist}
\end{figure*}

\subsubsection{Meta-training} We now describe the meta-training procedure, which is also presented in pseudocode in Algorithm \ref{alg:meta-training}. Formally, consider a base-learner with random initial weights $\theta$. We aim to learn good initialization weights for the base-learner, including the treatment embedding, through meta-iterations of meta-training. In each meta-iteration, we sample a training task $\tau\in T$  as well as $k$ training examples from each of the groups and train the model specifically for $\tau$. We denote the sampled training and testing data $D_{\tau}^{\text{train}\prime}$ and $D_{\tau}^{\text{fine-tune}\prime}$, respectively. We let $U^{k}(\theta, D)$ be the operator that updates $\theta$ after training on some dataset $D$ with $k$-shot samples, using any optimizer such as SGD or Adam. 

We first train the model on the sampled training data and obtain new weights $\tilde \theta=U^{k}(\theta, { D_\tau^{\text{train}\prime}})$. Using these new weights, we fine-tune the model on the sampled test data, which gives improved weights $\tilde \theta=U^{k}(\tilde{\theta},  D_\tau^{\text{test}\prime})$. Finally, at the end of the meta-iteration, we improve the initialization weights by updating $\theta$ with a fraction of the changes in $\tilde\theta$:
\begin{equation}\label{eq:reptile_update}
    \theta = \theta + \epsilon(\theta-\tilde\theta),
\end{equation}
where $\epsilon\in(0,1]$ is a meta step size that is linearly annealed to zero over the meta-iterations. In the next meta-iteration, $\theta$ is used to initialize another base-learner for another training task. For simplicity, we describe the situation where only one training task is selected per meta-iteration, but it is also possible to train on multiple tasks in one meta-iteration sequentially and update $\theta$  at the end of the meta-iteration.

\subsubsection{Meta-testing} The inference stage of the meta-learner is straightforward. We evaluate each task separately for each test task in $T_{\text{target}}$. Instead of sampling data, all available data is used. We initialize a base-learner with the current weights $\theta$ and train it on the training set to obtain updated weights $\tilde \theta=U_{ D_\tau^{\text{train}}}^{k}(\theta)$. The model is then directly deployed on the test set.

By virtue of the fact that the training tasks are associated with the target tasks, the existing treatment embeddings have already captured a good latent representation of the unseen (out-of-sample) treatment group for the target tasks. More importantly, as the model has been trained specifically on how other groups respond to the target task, it can transfer its knowledge to an unseen group.

\section{DATA}
We use three real-world human subjects datasets in this paper.\footnote{Due to governmental regulations, we are unable to release our datasets to the public.} Two of the datasets, SHARPS and PHITE, are randomized controlled trials that study the effects of interventions on fitness abilities. The third dataset, TILES, is an observational study that considers the effect of external atypical events on affect. In Figure \ref{fig:data_dist}, we plot the distributions of the target variables of the three datasets, separated by treatment groups. Below, we describe the details of each dataset and the target variables used for prediction.
\subsection{SHARP}
Strengthening Human Adaptive Reasoning and Problem Solving (SHARP) is a study of the effects of multimodal interventions on human fitness and cognitive abilities \cite{daugherty2018multi}. This study recruited undergraduate students for a 4-month long trial. Subjects were randomly placed in one of four conditions: (A) fitness training only ($N=53$); (B) fitness and cognitive training ($N=66$); (C) fitness, cognitive training, and meditation ($N=57$); and finally (D) active controls ($N=67$). Participants were given a battery of self-reported questionnaires as well as cognitive, physical, and diet assessments both before and after the treatment to examine their demographics, personality, cognitive ability, mental states, physical conditions, etc. Since our target variables are fitness-related, we describe the assessments related to fitness and diet in detail as follows:
\begin{itemize}
    \item Anthropometrics (e.g., height, weight, sex, BMI)
    \item Aerobic capacity (e.g., VO2) 
    \item DXA scans (e.g. bone density, body measurements)
    \item Accelerometer trackings to measure activeness
    \item Diet record 
    
\end{itemize}

The interventions were administered over 48 sessions. Group A participants received 48 sessions of fitness training. Groups B and C both received 28 sessions of fitness training and 20 sessions of other training (cognitive or meditation, respectively). To remove the effects of sex differences, we calculate residual features for any features that are significantly different via a 2-sample t-test between the male and the female at the $\alpha=0.05$ level. The sex-based residual features are calculated by taking the difference between the original feature values and the mean feature value of the people of the same sex. For assessments repeated both pre- and post-treatment, we calculate the difference in their post- and pre-treatment scores and use them as additional post-treatment features (this does \textit{not} include post-treatment target variables that we are predicting).

Our target variables are the differences between the post- and pre-treatment physical fitness assessment called the Army Physical Readiness Test (APRT), which measures a subject's ability to perform \textit{1.5 mile run}, \textit{long jump}, \textit{push-ups}, \textit{rowers}, and \textit{shuttle run}. The APRT test was administered pre- and post-intervention for participants in the first three groups. For this reason, we do not consider group D. We illustrate the distribution of the APRT differences for each group in Fig. \ref{fig:data_dist}(a). The target variables \textit{1.5 mile run} and \textit{shuttle run} have similar distributions across all groups, implying that the differences in treatments had minimal effect. But group A significantly improved in \textit{long jump}, \textit{push-ups}, and \textit{rowers}.

\subsection{PHITE}
Precision High-Intensity Training Through Epigenetics (PHITE) is a study that explores the link between physical fitness and epigenetics \cite{phite}. Of the 66 participants who completed the study, 33 each were randomly assigned to either a moderate or high-intensity 12-week exercise regimen as the treatment intervention.\footnote{We do not know which group received which treatment.} Both treatments consist of varying levels of endurance training and resistance training. Each participant was given a battery of cognitive tests, physical assessments, and biospecimen analyses before the treatment. Some assessments were repeated during or after the treatment. As with SHARP, we also calculate residual features that are distinctly different between males and females as well as differential features for applicable features taken at different time points.

We are interested in predicting the difference between post- and pre-treatment scores on three fitness assessments: the peak power achieved on a cycling machine, the number of pull-ups, and the lateral pulldown 1RM (one repetition maximum). The distributions of these fitness gains are shown in Fig.~\ref{fig:data_dist}(b).

\begin{figure*}
    \centering
    \includegraphics[width=1\linewidth]{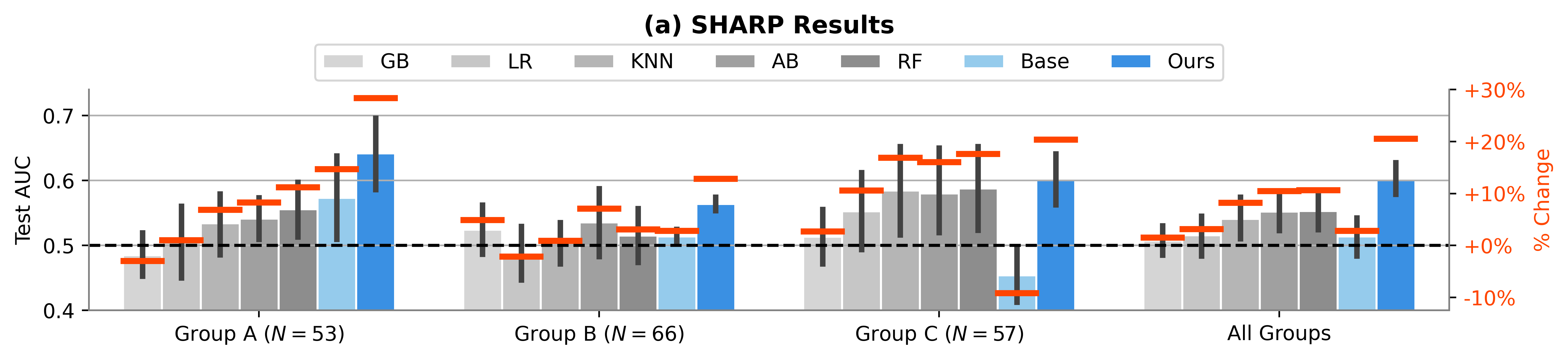}
    \includegraphics[width=0.85\linewidth]{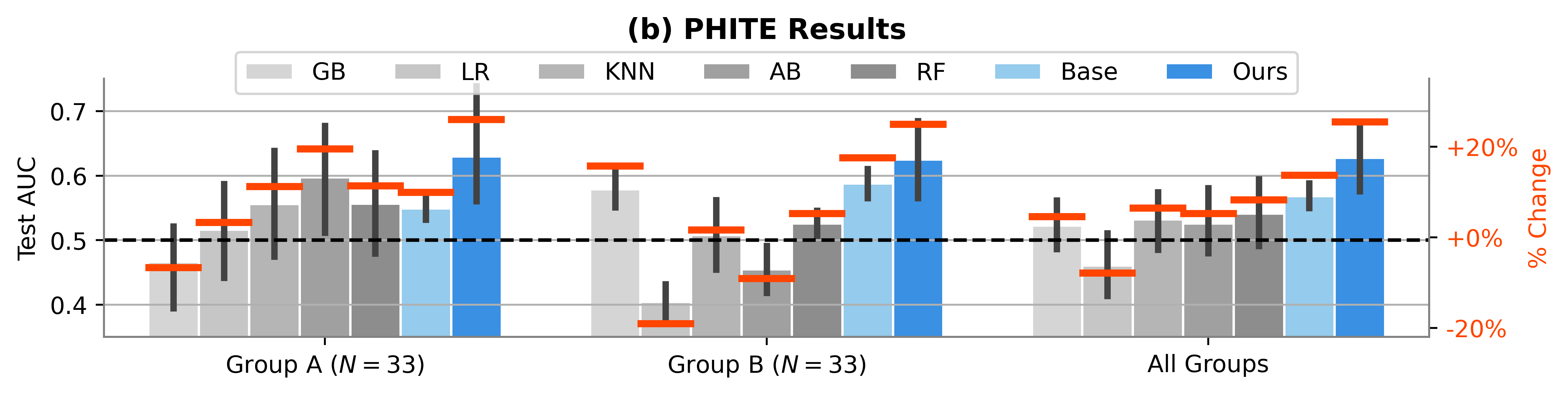}
    \caption{Classification results on the (a) SHARP and (b) PHITE datasets. The barplots correspond to the left-axes, which shows the average AUC (the higher the better) across all target variables. The error lines show the standard error. The grey bars are baseline methods, the light blue bars are the initial results from the base-learner, and the deep blue bars are the final results of our meta-learner. The red bars correspond to the right-axes, which show the percentage of change over an AUC of 0.50. }
    \label{fig:sharp_phite_res}
\end{figure*}

\subsection{TILES}
Tracking IndividuaL performancE with Sensors (TILES) is a multimodal longitudinal study that uses wearable sensors to track the daily activities of 212 hospital workers over 10 weeks~\cite{mundnich2020tiles}. All participants received a series of baseline surveys prior to the data collection process, which accessed their cognitive ability, physical ability, personality, sleep quality, and other behavioral assessments. Over the course of the study, each participant also received ecological momentary assessments (EMAs) twice daily, which briefly assessed their stress, anxiety, affect, job performance, and well-being that day. In particular, the EMAs asked if the participants had experienced any atypical events, for which they could answer in free-form texts. These texts are coded by human annotators.  Following \cite{burghardt2020having}, atypical events are one of \textit{minor positive}, \textit{minor negative}, \textit{major negative}, or \textit{neutral} (no atypical event). Similar to previous work, we use activity data given by the Fitbits they wore wore24 hours every day \cite{yan2019estimating,yan2020affect,burghardt2020having}. The Fitbits track their sleep, heart rates, steps, and physical activity. Every feature is aggregated per user daily.

Our goal is to predict the changes in the subject's mood (positive and negative affect) from their survey and sensor data, with their ``treatment groups'' determined by their atypical event categorization. Previous research suggests that these atypical events can strongly influence the subject's mood \cite{yan2020affect,burghardt2020having}. Additionally, atypical events are usually unexpected and short-termed, thus approximating a random assignment for any subject over time \cite{burghardt2020having}. A data point corresponds to a day for one individual, where the input features are the individual's features on that (target) day, the day before, and the day after, and the target variables are the changes in their positive and negative affect from the day before to the target day. Similar to SHARP and PHITE, we calculate differential values between the same types of features collected on consecutive days.

As participants exhibit various compliance rates \cite{yan2019estimating,yan2020affect,burghardt2020having}, we retain users with at least three days worth of consecutive data and users who have experienced at least one positive or negative atypical event. There are 127 eligible users totaling 4,371 entries of data points (4,013 \textit{neutral}, 103 \textit{minor positive}, 210 \textit{minor negative}, and 45 \textit{major negative}). Fig. \ref{fig:data_dist}(c) illustrates the distribution of the change in positive and negative affect, which are distinct for all groups. In particular, the \textit{major negative} event category experiences the most significant increase in negative affect.

\begin{figure*}
    \centering
    \includegraphics[width=\linewidth]{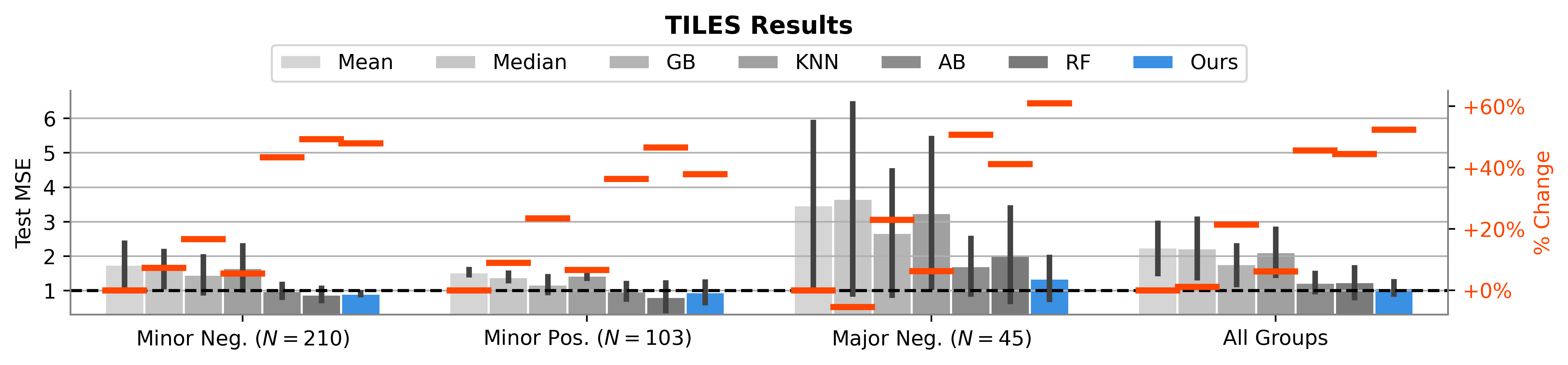}
    \caption{Regression results on the TILES dataset. The barplots correspond to the left-axes, which shows the average MSE (the lower the better) across the standardized target variables. The error lines show the standard error. The grey bars are baseline methods and the deep blue bars are the meta-learners. The red bars correspond to the right-axes, which show the percentage of change over the mean MSE predicted by the Mean method. }
    \label{fig:tiles_res}
\end{figure*}
\section{Results}

In order to assess our approach, we answer the following experimental questions (EQs) in this section: 
\begin{itemize}
    \item \textbf{EQ1}: Can our model enable zero-shot prediction for unseen treatment groups?
    \item \textbf{EQ2}: Does our model generalize well when the test group exhibits distinct distributions from the training group(s)?
    \item \textbf{EQ3}: Can model resist overfitting to the training groups? 
\end{itemize}

\subsection{Experiment Settings}
We evaluate our model via cross-validation by repeatedly splitting the training and testing data points by treatment groups. That is, in every split, one treatment group will be used as the test group and the rest as training groups. An exception is that the \textit{neutral} group for TILES is not made a test group, as there are minimal changes in affect for people who experience neutral events \cite{burghardt2020having}. 
For PHITE and SHARP, to facilitate the comparison across different target variables, we frame the prediction problems as binary classification problems and predict whether each target variable is greater than 0 (an increase in the corresponding fitness ability). For TILES, since there are many data points that exhibit significant increases or decreases in positive and negative affect, we frame the prediction tasks as regression problems. As all of these data contain missing features, we remove features that are largely missing (according to a threshold, see Appendix) or impute them using the mean values from the training set. The regression tasks are scored with MSE and the classification tasks are scored with AUC.

\subsubsection{Baselines} We use the following baselines: Random Forest (RF) \cite{randomforests}, AdaBoost (AB) \cite{adaboost}, GradientBoosting (GB) \cite{gradientboosting}, K-Nearest Neighbors (KNN) \cite{knn}, and Logistic Regression (LR) \cite{logisticregression} ((for classification only). For the regression problems, we additionally use the mean and median of the training set as baseline point estimates. We use the Scikit-Learn implementations of all baseline models and Tensorflow for the base- and meta-learner. All models are tuned via grid search. The details for the randomized grid searching process can be found in the Appendix.

\subsubsection{The Inapplicability of Meta-Learning Baselines} Our problem is distinct from those addressed in most ZSL literature, which is designed for zero-shot classification  \cite{romera2015embarrassingly,verma2020meta,xian2016latent,felix2018multi}, or meta-learning literature, which are designed for few-shot adaptation \cite{finn2017model, nichol2018reptile,li2016transfer}, research. To the best of our knowledge, there is no existing meta-learning algorithm that applies to our setting. As such, we cannot compare directly with these existing ZSL and meta-learning models.

\begin{figure*}
    \centering
    \includegraphics[width=\linewidth]{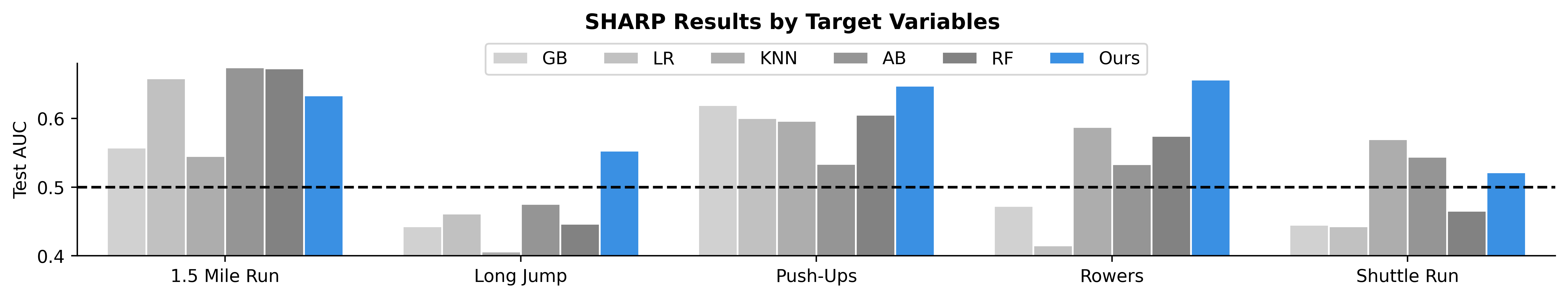}
    \caption{SHARP results by target variables show that the biggest improvements brought by our model are for the three target variables that have the largest variation between the treatment groups (long jump, push-up, and rowers) The bars indicate the average AUC of a model for the same target variable over all three groups.}
    \label{fig:sharp_res_by_y}
\end{figure*}
\begin{figure}
    \centering
    \includegraphics[width=\linewidth]{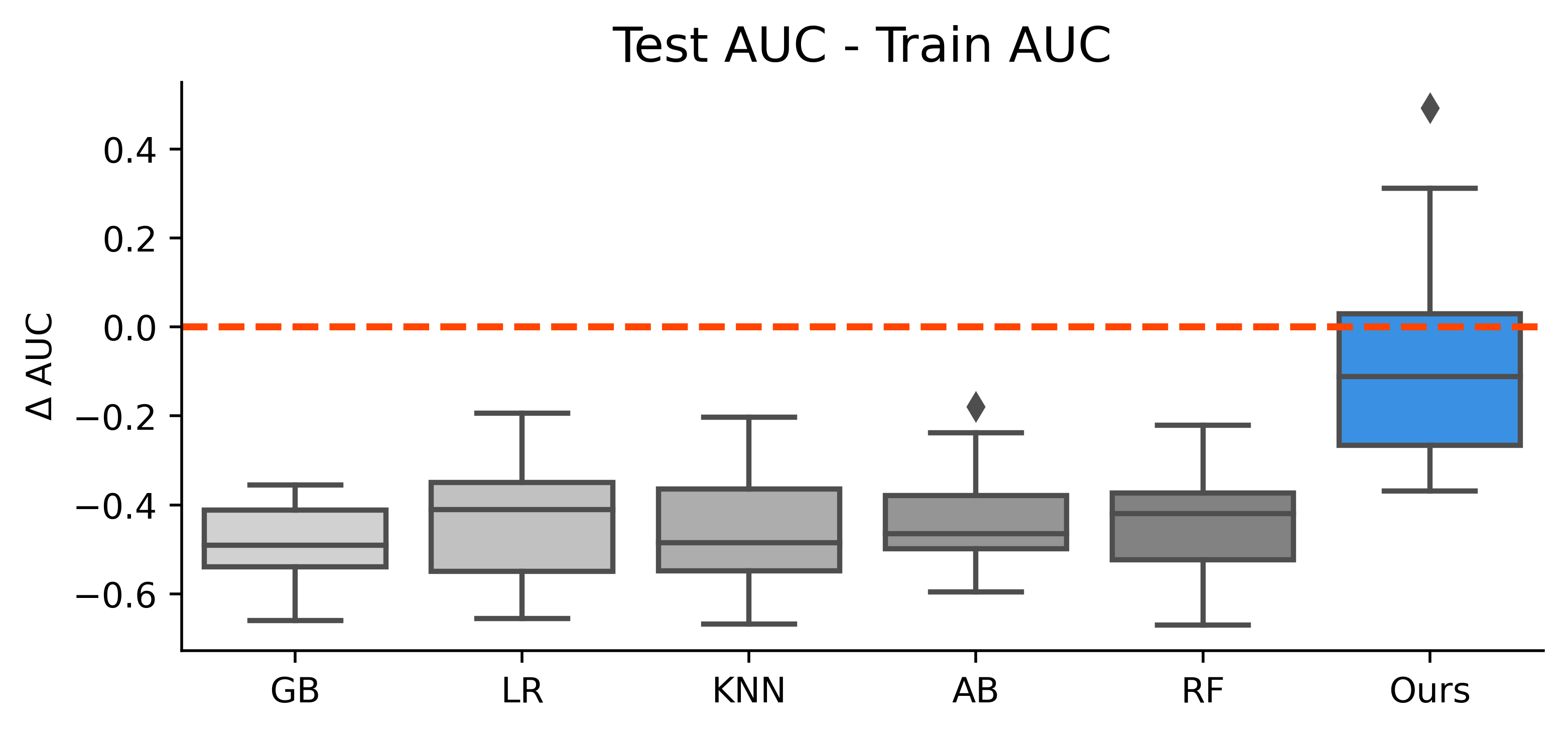}
    \caption{Boxplot distribution of the difference between the test AUCs and the training AUCs on the SHARP and PHITE datasets indicates that our model is the most resistant to overfitting to the training data.}
    \label{fig:train_test_auc_diff}
\end{figure}

\subsection{EQ1: Zero-shot Predictions}

Fig. \ref{fig:sharp_phite_res} shows the classification results on the SHARP and PHITE datasets. This figure shows that our model improves over baseline methods on both datasets and across all treatment groups. Our model produces the best results for each group individually and all groups combined. For all groups, we plot the percentage improvement of their AUCs over an AUC of 0.50, since an AUC of 0.50 is the same as random chance and an AUC any lower is worse than chance. For SHARP, the combined improvement in AUC is about 20\% and is close to 30\% specifically for group A. For PHITE, our model delivers an improvement of over 20\% for all groups. Moreover, we see that the meta-learning strategies significantly improve upon the base-learner, which alone can perform very poorly at times (less than 0.50 AUC).

The regression results for predicting changes in positive and negative affects on TILES are shown in Fig. \ref{fig:tiles_res}. We standardize both target variables with respect to the \textit{neutral} group. In predicting affect changes for the \textit{minor negative} and \textit{minor positive} groups, all models perform fairly well. Although the meta-learner is not the best-performing model, its MSEs are still within 1 standard deviation of the data. The meta-learner is best at predicting affect changes for the major negative group, with an improvement of over 60\% in MSE compared to the Mean method.

\subsection{EQ2: Shifts in Test Group Distribution}
We observe that our model does particularly well when the test groups' target variables have distinct distributions from the training groups, for instance in the case of group A in SHARP and the major negative group in TILES (Fig. \ref{fig:data_dist}). We investigate whether the distribution of training vs. test target variables plays a role in performance. In Fig. \ref{fig:sharp_res_by_y}, we plot the performance of all models by target variables. We see that for \textit{1.5 mile run} and \textit{shuttle run}, our model is not the best nor the worst performing model. The advantage of our model lies in predicting long jump, push-ups, and rowers, the same three target variables that have a significant difference in their distributions across the three groups (Fig. \ref{fig:data_dist}(a)). In particular, in predicting \textit{long jump}, all baseline models have an AUC lower than 0.50, indicating they do worse than random chance, whereas our model shows a significant improvement over an AUC of 0.50. Similarly, we find that the meta-learner is particularly good at predicting the change in negative affect for the \textit{major negative} group in TILES, which experienced the most dramatic increase in negative affect.
\subsection{EQ3: Resistance to Overfitting}
Since our model performs well in scenarios in which inter-group distributions are distinct, it suggests that it does not overfit the training distributions and, consequently, is most well-positioned to predict out-of-sample data. In Fig. \ref{fig:train_test_auc_diff}, we plot the difference in the testing and training AUCs. A difference of 0 indicates that the model does equally well on the test set and the training set. A difference of less than 1 suggests overfitting, and in traditional settings a difference greater than 1 suggests underfitting. However, as our challenge lies in transferring knowledge from the training to the test set,  we are less concerned with underfitting the training set and more concerned with the performance on the test set. From this figure, we see that all baseline models severely overfit the training set. In contrast, our model is resistant to overfitting, which also explains why our model performs particularly well for certain target variables and certain groups.

\section{Conclusion}
In this paper, we address the problem of predicting multiple treatment outcomes in a zero-shot scenario where the outcomes of treatment for an entire group of human subjects are unknown. Such problems are challenging because, by definition, the test data are in the out-out-sample. Our approach is an end-to-end framework based on meta-learning, which learns to learn good initialization weights for our target prediction tasks. By meta-learning from training tasks that are similar to the target tasks, our model learns good initialization weights that can adapt rapidly to new tasks. In this process, we also learn the embed the effects of each treatment group, including the held-out one, in the treatment representations. 

We evaluate our model on real-world small-scale data collected in human subject studies and show that our model outperforms baseline models in predicting outcomes for each group. In particular, we find that our model performs best when the training and testing distributions are most different because it is resistant to overfitting. Our model could be especially promising for generalizing from limited data on human subjects to a wider population.

Finally, we recognize that predicting with limited data is a challenging problem. The arguably more advanced deep learner architectures (identical to our base-learner models in Fig. \ref{fig:tiles_res} and Fig. \ref{fig:sharp_phite_res}) do not necessarily outperform traditional machine learning models (e.g., decision trees). With the meta-learning paradigm, however, we can vastly improve predictive performance with \textit{limited} and \textit{out-of-domain} without requiring more data.

\section*{Acknowledgments} 
The authors are grateful to DARPA for support under AIE Opportunity No DARPA-PA-18-02-07. This project does not necessarily reflect the position or policy of the Government; no official endorsement should be inferred.



\bibliographystyle{IEEEtran}
\bibliography{main}

\begin{thebibliography}{10}
\providecommand{\url}[1]{#1}
\csname url@samestyle\endcsname
\providecommand{\newblock}{\relax}
\providecommand{\bibinfo}[2]{#2}
\providecommand{\BIBentrySTDinterwordspacing}{\spaceskip=0pt\relax}
\providecommand{\BIBentryALTinterwordstretchfactor}{4}
\providecommand{\BIBentryALTinterwordspacing}{\spaceskip=\fontdimen2\font plus
\BIBentryALTinterwordstretchfactor\fontdimen3\font minus
  \fontdimen4\font\relax}
\providecommand{\BIBforeignlanguage}[2]{{%
\expandafter\ifx\csname l@#1\endcsname\relax
\typeout{** WARNING: IEEEtran.bst: No hyphenation pattern has been}%
\typeout{** loaded for the language `#1'. Using the pattern for}%
\typeout{** the default language instead.}%
\else
\language=\csname l@#1\endcsname
\fi
#2}}
\providecommand{\BIBdecl}{\relax}
\BIBdecl

\bibitem{ildstad2001small}
S.~T. Ildstad, C.~H. Evans~Jr \emph{et~al.}, \emph{Small clinical trials:
  issues and challenges}.\hskip 1em plus 0.5em minus 0.4em\relax National
  Academies Press, 2001.

\bibitem{hamburg2010path}
M.~A. Hamburg and F.~S. Collins, ``The path to personalized medicine,''
  \emph{NEJM}, vol. 363, no.~4, pp. 301--304, 2010.

\bibitem{day2018recommendations}
S.~Day, A.~H. Jonker, L.~P.~L. Lau, R.-D. Hilgers, I.~Irony, K.~Larsson, K.~C.
  Roes, and N.~Stallard, ``Recommendations for the design of small population
  clinical trials,'' \emph{Orphanet J. Rare Dis.}, vol.~13, no.~1, p. 195,
  2018.

\bibitem{hackshaw2008small}
A.~Hackshaw, ``Small studies: strengths and limitations,'' \emph{European Resp.
  J.}, vol.~32, no.~5, pp. 1141--1143, 2008.

\bibitem{martin2017much}
L.~Martin, M.~Hutchens, C.~Hawkins, and A.~Radnov, ``How much do clinical
  trials cost?'' \emph{Nature Rev. Drug Discov.}, vol.~16, pp. 381--382, 2017.

\bibitem{davis1994generalizing}
C.~Davis, ``Generalizing from clinical trials,'' \emph{Controlled Clinical
  Trials}, vol.~15, no.~1, pp. 11--14, 1994.

\bibitem{flather2006generalizing}
M.~Flather, N.~Delahunty, and J.~Collinson, ``Generalizing results of
  randomized trials to clinical practice: reliability and cautions,''
  \emph{Clinical Trials}, vol.~3, no.~6, pp. 508--512, 2006.

\bibitem{hong2017generalizing}
J.-L. Hong, M.~Jonsson~Funk, R.~LoCasale, S.~E. Dempster, S.~R. Cole,
  M.~Webster-Clark, J.~K. Edwards, and T.~Stürmer, ``Generalizing randomized
  clinical trial results: Implementation and challenges related to missing data
  in the target population,'' \emph{American J. Epidemiology}, vol. 187, no.~4,
  pp. 817--827, 08 2017.

\bibitem{voulodimos2018deep}
A.~Voulodimos, N.~Doulamis, A.~Doulamis, and E.~Protopapadakis, ``Deep learning
  for computer vision: A brief review,'' \emph{Comput. Intell. Neurosci.}, vol.
  2018, 2018.

\bibitem{devlin2018bert}
J.~Devlin, M.-W. Chang, K.~Lee, and K.~Toutanova, ``{BERT}: Pre-training of
  deep bidirectional transformers for language understanding,'' in
  \emph{ACL-HLT '19}.\hskip 1em plus 0.5em minus 0.4em\relax ACL, 2019, pp.
  4171--4186.

\bibitem{rajkomar2018scalable}
A.~Rajkomar, E.~Oren, K.~Chen, A.~M. Dai, N.~Hajaj, M.~Hardt, P.~J. Liu,
  X.~Liu, J.~Marcus, M.~Sun \emph{et~al.}, ``Scalable and accurate deep
  learning with electronic health records,'' \emph{NPJ Digital Med.}, vol.~1,
  no.~1, p.~18, 2018.

\bibitem{chekroud2016cross}
A.~M. Chekroud, R.~J. Zotti, Z.~Shehzad, R.~Gueorguieva, M.~K. Johnson, M.~H.
  Trivedi, T.~D. Cannon, J.~H. Krystal, and P.~R. Corlett, ``Cross-trial
  prediction of treatment outcome in depression: a machine learning approach,''
  \emph{The Lancet Psychiatry}, vol.~3, no.~3, pp. 243--250, 2016.

\bibitem{walsh2017predicting}
C.~G. Walsh, J.~D. Ribeiro, and J.~C. Franklin, ``Predicting risk of suicide
  attempts over time through machine learning,'' \emph{Clinical Psychological
  Sci.}, vol.~5, no.~3, pp. 457--469, 2017.

\bibitem{sun2017revisiting}
C.~{Sun}, A.~{Shrivastava}, S.~{Singh}, and A.~{Gupta}, ``Revisiting
  unreasonable effectiveness of data in deep learning era,'' in \emph{ICCV
  '17}.\hskip 1em plus 0.5em minus 0.4em\relax IEEE, 2017, pp. 843--852.

\bibitem{marcus2018deep}
G.~Marcus, ``Deep learning: A critical appraisal,'' \emph{arXiv preprint
  arXiv:1801.00631}, 2018.

\bibitem{shah2019artificial}
P.~Shah, F.~Kendall, S.~Khozin, R.~Goosen, J.~Hu, J.~Laramie, M.~Ringel, and
  N.~Schork, ``Artificial intelligence and machine learning in clinical
  development: a translational perspective,'' \emph{NPJ Digital Med.}, vol.~2,
  no.~1, pp. 1--5, 2019.

\bibitem{li2016transfer}
Y.~Li, L.~Wang, J.~Wang, J.~Ye, and C.~K. Reddy, ``Transfer learning for
  survival analysis via efficient {L2}, 1-norm regularized cox regression,'' in
  \emph{ICDM '16}.\hskip 1em plus 0.5em minus 0.4em\relax IEEE, 2016, pp.
  231--240.

\bibitem{qiu2020meta}
Y.~L. Qiu, H.~Zheng, A.~Devos, H.~Selby, and O.~Gevaert, ``A meta-learning
  approach for genomic survival analysis,'' \emph{Nature Commun.}, vol.~11,
  no.~1, pp. 1--11, 2020.

\bibitem{finn2017model}
C.~Finn, P.~Abbeel, and S.~Levine, ``Model-agnostic meta-learning for fast
  adaptation of deep networks,'' in \emph{ICML '17}.\hskip 1em plus 0.5em minus
  0.4em\relax PMLR, 2017, pp. 1126--1135.

\bibitem{romera2015embarrassingly}
B.~Romera-Paredes and P.~Torr, ``An embarrassingly simple approach to zero-shot
  learning,'' in \emph{ICML '15}.\hskip 1em plus 0.5em minus 0.4em\relax PMLR,
  2015, pp. 2152--2161.

\bibitem{verma2020meta}
V.~K. Verma, D.~Brahma, and P.~Rai, ``Meta-learning for generalized zero-shot
  learning,'' in \emph{AAAI '20}, vol.~34, no.~04, 2020, pp. 6062--6069.

\bibitem{xian2016latent}
Y.~Xian, Z.~Akata, G.~Sharma, Q.~Nguyen, M.~Hein, and B.~Schiele, ``Latent
  embeddings for zero-shot classification,'' in \emph{Proceedings of the IEEE
  conference on computer vision and pattern recognition}, 2016, pp. 69--77.

\bibitem{felix2018multi}
R.~Felix, I.~Reid, G.~Carneiro \emph{et~al.}, ``Multi-modal cycle-consistent
  generalized zero-shot learning,'' in \emph{Proceedings of the European
  Conference on Computer Vision (ECCV)}, 2018, pp. 21--37.

\bibitem{nichol2018reptile}
A.~Nichol and J.~Schulman, ``On first-order meta-learning algorithms,''
  \emph{arXiv preprint arXiv:1803.02999}, 2018.

\bibitem{pan2010survey}
S.~J. Pan and Q.~Yang, ``A survey on transfer learning,'' \emph{IEEE TKDE},
  vol.~22, no.~10, pp. 1345--1359, 2010.

\bibitem{csurka2017domain}
G.~Csurka, \emph{Domain adaptation in computer vision applications}.\hskip 1em
  plus 0.5em minus 0.4em\relax Springer, 2017.

\bibitem{ruder2019transfer}
S.~Ruder, M.~E. Peters, S.~Swayamdipta, and T.~Wolf, ``Transfer learning in
  natural language processing,'' in \emph{NAACL-HLT '19}, 2019, pp. 15--18.

\bibitem{long2017deep}
M.~Long, H.~Zhu, J.~Wang, and M.~I. Jordan, ``Deep transfer learning with joint
  adaptation networks,'' in \emph{ICML '17}.\hskip 1em plus 0.5em minus
  0.4em\relax PMLR, 2017, pp. 2208--2217.

\bibitem{tzeng2017adversarial}
E.~Tzeng, J.~Hoffman, K.~Saenko, and T.~Darrell, ``Adversarial discriminative
  domain adaptation,'' in \emph{CVPR '17}.\hskip 1em plus 0.5em minus
  0.4em\relax IEEE, July 2017.

\bibitem{zhu2020ensemble}
Y.~Zhu, T.~Brettin, Y.~A. Evrard, A.~Partin, F.~Xia, M.~Shukla, H.~Yoo, J.~H.
  Doroshow, and R.~L. Stevens, ``Ensemble transfer learning for the prediction
  of anti-cancer drug response,'' \emph{Sci. Rep.}, vol.~10, no.~1, pp. 1--11,
  2020.

\bibitem{caruana1997multitask}
R.~Caruana, ``Multitask learning,'' \emph{Mach. Learn.}, vol.~28, no.~1, pp.
  41--75, 1997.

\bibitem{sousa2006task}
J.~P. Sousa, V.~Poladian, D.~Garlan, B.~Schmerl, and M.~Shaw, ``Task-based
  adaptation for ubiquitous computing,'' \emph{IEEE Trans. Syst., Man, Cybern.
  C. Appl. Rev.}, vol.~36, no.~3, pp. 328--340, 2006.

\bibitem{zhang2017survey}
Y.~Zhang and Q.~Yang, ``A survey on multi-task learning,'' \emph{arXiv preprint
  arXiv:1707.08114}, 2017.

\bibitem{liu2019multitask}
X.~Liu, P.~He, W.~Chen, and J.~Gao, ``Multi-task deep neural networks for
  natural language understanding,'' in \emph{ACL '19}.\hskip 1em plus 0.5em
  minus 0.4em\relax ACL, Jul. 2019, pp. 4487--4496.

\bibitem{torrey2010transfer}
L.~Torrey and J.~Shavlik, ``Transfer learning,'' in \emph{Handbook of research
  on machine learning applications and trends: algorithms, methods, and
  techniques}.\hskip 1em plus 0.5em minus 0.4em\relax IGI Global, 2010, pp.
  242--264.

\bibitem{taylor2009transfer}
M.~E. Taylor and P.~Stone, ``Transfer learning for reinforcement learning
  domains: A survey.'' \emph{JMLR}, vol.~10, no.~7, 2009.

\bibitem{yogatama2019learning}
D.~Yogatama, C.~d.~M. d'Autume, J.~Connor, T.~Kocisky, M.~Chrzanowski, L.~Kong,
  A.~Lazaridou, W.~Ling, L.~Yu, C.~Dyer \emph{et~al.}, ``Learning and
  evaluating general linguistic intelligence,'' \emph{arXiv preprint
  arXiv:1901.11373}, 2019.

\bibitem{vu2020exploring}
T.~Vu, T.~Wang, T.~Munkhdalai, A.~Sordoni, A.~Trischler, A.~Mattarella-Micke,
  S.~Maji, and M.~Iyyer, ``Exploring and predicting transferability across
  {NLP} tasks,'' \emph{arXiv preprint arXiv:2005.00770}, 2020.

\bibitem{li2020transfer}
X.~Li, Y.~Grandvalet, F.~Davoine, J.~Cheng, Y.~Cui, H.~Zhang, S.~Belongie,
  Y.-H. Tsai, and M.-H. Yang, ``Transfer learning in computer vision tasks:
  Remember where you come from,'' \emph{Image Vision Comput.}, vol.~93, p.
  103853, 2020.

\bibitem{wang2020generalizing}
Y.~Wang, Q.~Yao, J.~T. Kwok, and L.~M. Ni, ``Generalizing from a few examples:
  A survey on few-shot learning,'' \emph{ACM Comput. Surv.}, vol.~53, no.~3,
  Jun. 2020.

\bibitem{ravi2016optimization}
S.~Ravi and H.~Larochelle, ``Optimization as a model for few-shot learning,''
  in \emph{ICLR '16}.\hskip 1em plus 0.5em minus 0.4em\relax OpenReview.net,
  2016.

\bibitem{vinyals2016matching}
O.~Vinyals, C.~Blundell, T.~Lillicrap, D.~Wierstra \emph{et~al.}, ``Matching
  networks for one shot learning,'' in \emph{NIPS '16}, vol.~29.\hskip 1em plus
  0.5em minus 0.4em\relax Curran Associates, Inc., 2016, pp. 3630--3638.

\bibitem{snell2017prototypical}
J.~Snell, K.~Swersky, and R.~Zemel, ``Prototypical networks for few-shot
  learning,'' in \emph{NIPS '17}.\hskip 1em plus 0.5em minus 0.4em\relax Curran
  Associates, Inc., 2017, pp. 4077--4087.

\bibitem{santoro2016meta}
A.~Santoro, S.~Bartunov, M.~Botvinick, D.~Wierstra, and T.~Lillicrap,
  ``Meta-learning with memory-augmented neural networks,'' in \emph{ICML
  '16}.\hskip 1em plus 0.5em minus 0.4em\relax PMLR, 2016, pp. 1842--1850.

\bibitem{li2017meta}
Z.~Li, F.~Zhou, F.~Chen, and H.~Li, ``Meta-{SGD}: Learning to learn quickly for
  few-shot learning,'' \emph{arXiv preprint arXiv:1707.09835}, 2017.

\bibitem{kunzel2019metalearners}
S.~R. K{\"u}nzel, J.~S. Sekhon, P.~J. Bickel, and B.~Yu, ``Metalearners for
  estimating heterogeneous treatment effects using machine learning,''
  \emph{PNAS}, vol. 116, no.~10, pp. 4156--4165, 2019.

\bibitem{kraskov2004estimating}
A.~Kraskov, H.~St{\"o}gbauer, and P.~Grassberger, ``Estimating mutual
  information,'' \emph{Phys. Rev. E}, vol.~69, no.~6, p. 066138, 2004.

\bibitem{srivastava2014dropout}
N.~Srivastava, G.~Hinton, A.~Krizhevsky, I.~Sutskever, and R.~Salakhutdinov,
  ``Dropout: a simple way to prevent neural networks from overfitting,''
  \emph{JMLR}, vol.~15, no.~1, pp. 1929--1958, 2014.

\bibitem{salimans2016weight}
T.~Salimans and D.~P. Kingma, ``Weight normalization: A simple
  reparameterization to accelerate training of deep neural networks,'' in
  \emph{NIPS '16}.\hskip 1em plus 0.5em minus 0.4em\relax Curran Associates,
  Inc., 2016, p. 901–909.

\bibitem{daugherty2018multi}
A.~M. Daugherty, C.~Zwilling, E.~J. Paul, N.~Sherepa, C.~Allen, A.~F. Kramer,
  C.~H. Hillman, N.~J. Cohen, and A.~K. Barbey, ``Multi-modal fitness and
  cognitive training to enhance fluid intelligence,'' \emph{Intell.}, vol.~66,
  pp. 32--43, 2018.

\bibitem{phite}
\BIBentryALTinterwordspacing
ClinicalTrials.gov, ``Precision high-intensity training through epigenetics
  ({PHITE}),'' 2017-2021, identifier NCT03380923. [Online]. Available:
  \url{https://clinicaltrials.gov/ct2/show/NCT03380923}
\BIBentrySTDinterwordspacing

\bibitem{mundnich2020tiles}
K.~Mundnich, B.~M. Booth, M.~l'Hommedieu, T.~Feng, B.~Girault, J.~L'Hommedieu,
  M.~Wildman, S.~Skaaden, A.~Nadarajan, J.~L. Villatte, T.~H. Falf, K.~Lerman,
  E.~Ferrara, and S.~Narayanan, ``{TILES}-2018: A longitudinal physiologic and
  behavioral data set of hospital workers,'' \emph{Sci. Data}, vol.~7, p. 354,
  2020.

\bibitem{burghardt2020having}
K.~Burghardt, N.~Tavabi, E.~Ferrara, S.~Narayanan, and K.~Lerman, ``Having a
  bad day? {D}etecting the impact of atypical life events using wearable
  sensors,'' \emph{arXiv preprint arXiv:2008.01723}, 2020.

\bibitem{yan2019estimating}
S.~Yan, H.~Hosseinmardi, H.-T. Kao, S.~Narayanan, K.~Lerman, and E.~Ferrara,
  ``Estimating individualized daily self-reported affect with wearable
  sensors,'' in \emph{ICHI '19}.\hskip 1em plus 0.5em minus 0.4em\relax IEEE,
  2019, pp. 1--9.

\bibitem{yan2020affect}
------, ``Affect estimation with wearable sensors,'' \emph{J. Healthc. Inform.
  Res.}, pp. 1--34, 2020.

\bibitem{randomforests}
L.~Breiman, ``Random forests,'' \emph{Mach. Learn.}, vol.~45, no.~1, pp. 5--32,
  2001.

\bibitem{adaboost}
Y.~Freund, R.~E. Schapire \emph{et~al.}, ``Experiments with a new boosting
  algorithm,'' in \emph{ICML '96}, vol.~96.\hskip 1em plus 0.5em minus
  0.4em\relax Citeseer, 1996, pp. 148--156.

\bibitem{gradientboosting}
J.~H. Friedman, ``Greedy function approximation: a gradient boosting machine,''
  \emph{Ann. Stat.}, pp. 1189--1232, 2001.

\bibitem{knn}
C.~M. Bishop, \emph{Pattern recognition and machine learning}.\hskip 1em plus
  0.5em minus 0.4em\relax springer, 2006.

\bibitem{logisticregression}
P.~McCullagh and J.~A. Nelder, \emph{Generalized linear models}.\hskip 1em plus
  0.5em minus 0.4em\relax Routledge, 2019.

\end{thebibliography}
\appendix
\section{Model selection} \label{appendix-data}

For all models, we explore various feature scaling (normalize, standardize, or as-is) methods and the threshold for removing missing feature values. We use randomized grid search for hyperparameter selection for all models, including the baselines. The best models are selected via cross-validation over all test groups and all target variables. We also record the initial results of the base-learner (\S\ref{sec:base-learner}) prior to meta-training to study the effects of the meta-learning.

For our meta-learning model, we experimented with various training task selection methods, base-learner architecture, and meta-learning settings. The training tasks can be all post-treatment features or the top 70\% to 99\% of features according to Pearson correlation or Mutual Information. For the base-learner, we varied the number of layers $\{2, 4, 6, 8\}$, dimension of treatment embeddings $\{8, 16, 32, 64, 128\}$, size of hidden units $\{8, 16, 32, 64, 128\}$, activation functions (ReLU or TanH), optimizer (Adam or SGD), number of dropout units $\{0.05, 0.1, 0.2\}$, regularization types (L1, L2, or both), regularization strengths $\{0.01, 0.001, ..., 0.000001\}$, and number of iterations $\{1, 2, 5\}$. For the meta-learner, we varied the number of meta-iterations (20 to 100, in increments of 10), meta step size $\{0.25, 0.50, 0.75\}$, number of training data sampled per group $\{5, 10, 15\}$, and number of training tasks per meta-iteration $\{1, 2, 5\}$. 
\end{document}